\documentclass[runningheads]{llncs}

\usepackage{eccv}

\usepackage{eccvabbrv}

\usepackage{graphicx}
\usepackage{booktabs}

\usepackage[accsupp]{axessibility}  

\usepackage{hyperref}

\usepackage{orcidlink}
\usepackage{graphicx}
\usepackage{multirow}
\usepackage{tikz}
\usepackage{comment}
\usepackage{amsmath,amssymb} 
\usepackage{color}
\usepackage{wrapfig} 
\usepackage{booktabs}
\usepackage{multirow}
\usepackage{amssymb}
\usepackage{bbding}

\begin{document}

\title{Centering the Value of Every Modality: Towards Efficient and Resilient Modality-agnostic Semantic Segmentation} 
\titlerunning{Towards Efficient and Resilient Modality-agnostic Semantic Segmentation}

\author{Xu Zheng\inst{1}\orcidlink{0000-0003-4008-8951} \and
Yuanhuiyi Lyu\inst{1}\orcidlink{0009-0004-1450-811X} \and
Jiazhou Zhou\inst{1}\orcidlink{0009-0008-5258-1675}
\and
Lin Wang\inst{1,2}\orcidlink{0000-0002-7485-4493}\thanks{Corresponding author}}

\authorrunning{X. Zheng et al.}

\institute{Hong Kong University of Science and Technology, Guangzhou, China \\
\email{zhengxu128@gmail.com, \{yuanhuiyilv, jiazhouzhou\}@hkust-gz.edu.cn}
\and
Hong Kong University of Science and Technology, Hong Kong, China \\
\email{linwang@ust.hk}\\
\url{https://vlislab22.github.io/MAGIC/} 
}

\maketitle

\begin{abstract}
Fusing an arbitrary number of modalities is vital for achieving robust multi-modal fusion of semantic segmentation yet remains less explored to date. Recent endeavors regard RGB modality as the center and the others as the auxiliary, yielding an asymmetric architecture with two branches. However, the RGB modality may struggle in certain circumstances, \eg, nighttime, while others, \eg, event data, own their merits; thus, it is imperative for the fusion model to discern robust and fragile modalities, and incorporate the most robust and fragile ones to learn a resilient multi-modal framework. To this end,  
we propose a novel method, named \textbf{MAGIC}, that can be flexibly paired with various backbones, ranging from compact to high-performance models. Our method comprises two key plug-and-play modules. Firstly, we introduce a multi-modal aggregation module to efficiently process features from multi-modal batches and extract complementary scene information. On top, a unified arbitrary-modal selection module is proposed to utilize the aggregated features as the benchmark to rank the multi-modal features based on the similarity scores. This way, our method can eliminate the dependence on RGB modality and better overcome sensor failures while ensuring the segmentation performance. Under the commonly considered multi-modal setting, our method achieves state-of-the-art performance while reducing the model parameters by \textbf{60\%}. Moreover, our method is superior in the novel modality-agnostic setting, where it outperforms prior arts by a large margin of \textbf{+19.41}\% mIoU.


\end{abstract}

\begin{figure}[t!]
    \centering
    \includegraphics[width=\linewidth]{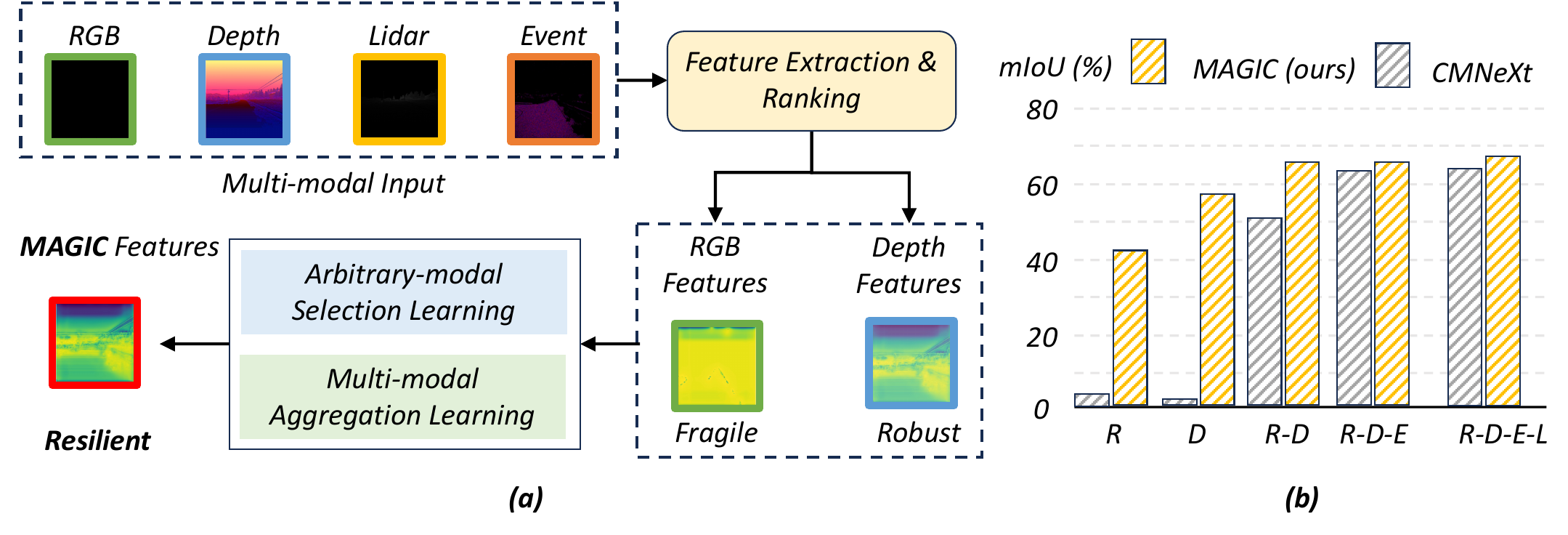}
    \caption{ 
    (a): Robust and fragile features extracted from multi-modal data and the resilient one learned by MAGIC.  
    (b): Modality-agnostic segmentation with arbitrary inputs using \{\textbf{R}GB, \textbf{D}epth, \textbf{E}vent, \textbf{L}iDAR\} on DELIVER. 
    }
    \label{cover_fig}
\end{figure}

\section{Introduction}
Nature has elucidated that miscellaneous sense and processing capabilities of visual information are vital for the understanding of the sophisticated environment~\cite{duan2022multimodal,su2023recent}. 
As such, advanced robots or self-driving vehicles need multi-sensor systems, which encompass diverse sensors, such as RGB, LiDAR, and event cameras, to achieve reliable scene understanding, including semantic segmentation~\cite{huang2022multi, wang2023multi, milioto2019rangenet++, zhang2020polarnet, alonso2019ev, jia2023event, sun2019rtfnet, wang2016learning, park2017rdfnet}. The intuition is that each sensor delivers its distinct advantages~\cite{wang2021survey, zheng2023deep}. 

Early endeavors predominantly focus on developing on bespoke fusion methods for specific sensors (\eg, RGB-depth~\cite{song2022improving}, RGB-Lidar~\cite{li2023mseg3d}, RGB-event~\cite{zhang2021issafe}, RGB-thermal~\cite{hui2023bridging}). Unfortunately, these methods lack flexibility and versatility when incorporating additional sensors.
As a result, fusing an arbitrary number of modalities is trendily desirable for achieving more robust multi-modal fusion in semantic segmentation. 
Despite its importance, this research direction remains relatively unexplored.
Only recently, a few works have been proposed, which regard the RGB modality as the primary while others as the auxiliary~\cite{zhu2023visual,zhang2022cmx,zhang2023delivering}. 
Naturally, a unified RGB-X pipeline is developed, yielding a distributed joint branch or an asymmetric architecture featuring two branches. 
In particular, CMNeXt~\cite{zhang2023delivering} introduces a self-query hub to extract effective information from any auxiliary modalities for subsequent fusion with the RGB modality. 

\textbf{Motivation:} Nevertheless, the RGB modality may struggle in certain circumstances, as demonstrated by the visualized feature in nighttime condition in  Fig.~\ref{cover_fig} (a). By contrast, alternative sensors offer distinct benefits that bolster scene understanding in demanding settings. 
For instance, depth cameras, with their ability to function reliably in low-light conditions and provide spatial data independent of ambient lighting, are valuable in the nighttime applications.
This indicates that \textbf{\textit{only by centering the value of every modality}},
we can essentially harness the superiority of all modalities for achieving modality-agnostic segmentation. Therefore, it becomes imperative for the fusion model to distinguish the \textbf{robust} and \textbf{fragile} modalities, and subsequently incorporate the most robust and fragile modalities to learn a more \textbf{resilient} multi-modal framework. The most robust feature is used to \textbf{enhance the segmentation accuracy} while the most fragile feature is incorporated to \textbf{reinforce the framework's resilience} against missing modalities.

\textbf{Contributions:} In light of this, we propose an efficient and robust \textbf{M}odality-\textbf{ag}nost\textbf{ic} (\textbf{MAGIC}) segmentation framework, that can be flexibly paired with various backbones, ranging from efficient to high-performance models.
Our method comprises two plug-and-play modules that enable efficient multi-modal learning and improve the modality-agnostic robustness of segmentation models.
Firstly, we introduce a Multi-modal Aggregation Module (MAM) to efficiently process features from multi-modal data and extract complementary scene information from all the modalities without relying on a specific one.

On top, an Arbitrary-modal Selection Module (ASM) is proposed to dynamically fuse the modality-agnostic scene features during training and improve the backbone model's robustness with arbitrary-modal input during inference. Specifically,
it utilizes the aggregated features from MAM as a benchmark to rank the multi-modal features based on the similarity scores. It then merges the selected salient features together to obtain predictions, so as to achieve modality-agnostic capabilities.
This way, we can eradicate the reliance on RGB modality and better overcome sensor failures while significantly enhancing the segmentation performance, see Fig.~\ref{cover_fig} (b) and (c).
Moreover, our method combines the MAM's prediction and the ground truth to soften the supervision for the predictions of ASM for better convergence and to prevent training instability.

Extensive experiments under the widely considered multi-modal segmentation setting on two multi-modal benchmarks show that our framework outperforms the existing multi-modal semantic segmentation methods (\textbf{+1.33\%} \& \textbf{+1.47\%})
while reducing model parameters by \textbf{60\%}.
Moreover, we evaluate our method in the novel modality-agnostic settings with arbitrary-modal inputs. The results show that our methods significantly outperforms existing works by a large margin (\textbf{+19.41\%} \& \textbf{+12.83\%}).

\section{Related Work}
\noindent \textbf{Semantic Segmentation} is a fundamental vision task with many applications, such as autonomous driving~\cite{zheng2023distilling,zheng2024semantics,chen2024frozen,chen2023clip,zheng2023both,zheng2023look,zhu2023good,zheng2022transformer,feng2020deep, siam2018comparative, muhammad2022vision, wang2022sfnet, li2022self, xiao2023baseg, fantauzzo2022feddrive, nesti2022evaluating, cheng2022cenet}. Approaches for semantic segmentation can be categorized according to their basic computing paradigm, namely convolution and self-attention. The fully convolutional networks (FCN)\cite{long2015fully} made significant progress in semantic segmentation with their end-to-end pixel-wise classification paradigm. Subsequent works improve the performance by exploring the multi-scale features~\cite{chen2017deeplab, chen2018encoder, hou2020strip, zhao2017pyramid}, attention blocks~\cite{choi2020cars, fu2019dual, huang2019ccnet, yuan2021ocnet}, edge cues~\cite{borse2021inverseform, ding2019boundary, gong2021boundary, li2020improving, takikawa2019gated}, and context priors~\cite{hu2019acnet, lin2017refinenet, yu2020context, zhang2018context}. More recently, various self-attention-based transformers have been proposed for semantic segmentation~\cite{strudel2021segmenter, zheng2021rethinking, xie2021segformer, gu2022multi, zhang2022topformer, zhu2021unified, wang2022rtformer, xu2022multi, zhang2022segvit, liu2021swin, liu2022swin, wang2021pyramid, wang2022pvt}. 
 
While these works achieve promising performance on RGB images under perfect lighting and motion conditions, they still suffer under extreme scenarios with complex lighting and weather conditions. We incorporate these segmentation models as the backbones, and propose two plug-and-play modules to achieve robust multi-modality semantic segmentation and modality-agnostic segmentation with arbitrary-modal input data.

\noindent \textbf{Multi-modal Semantic Segmentation} has been extensively studied, with a focus on fusing the RGB modality with complementary modalities such as depth~\cite{lyu2024omnibind,lyu2024unibind,lyu2024image,wang2020learning, zhou2020rgb, wang2020deep, cao2021shapeconv, chen2021spatial, ying2022uctnet, lee2022spsn, cong2022cir, ji2022dmra, wang2022learning, song2022improving}, thermal~\cite{shivakumar2020pst900, zhang2021abmdrnet, wu2022complementarity, liao2022cross, zhou2023mmsmcnet, xie2023cross, chen2022modality, pang2023caver, hui2023bridging, zhang2023efficient}, polarization~\cite{kalra2020deep, mei2022glass, xiang2021polarization}, events~\cite{alonso2019ev, zhang2021issafe,zheng2024eventdance,cao2023chasing,zheng2024eventdance,zhou2024exact}, and LiDAR~\cite{zhuang2021perception, yan20222dpass, wang2022multimodal, li2022deepfusion, borse2023x, zhang2023mx2m, liu2022camliflow, li2023mseg3d}. With the development of novel sensors, various approaches~\cite{zheng2024360sfuda++,wang2022multimodal, wang2020deep, wang2020learning, zhang2022cmx, liang2022multimodal, zhang2023delivering} have been proposed to scale from dual modality fusion to multiple modality fusion for robust scene understanding abilities, \eg, MCubeSNet~\cite{liang2022multimodal}.

From the architecture design perspective, these methods can be divided into three categories, including merging with separate branches~\cite{broedermann2022hrfuser, wei2023mmanet, zhang2021abmdrnet, man2023bev}, distributing with a joint branch~\cite{wang2022multimodal, chen2021spatial}, and fusion with asymmetric branches~\cite{zhang2023delivering, zhang2022cmx}. These approaches regard RGB modality as the center and the other modalities as the auxiliary, yielding an asymmetric architecture with two branches. In particular, CMNeXT~\cite{zhang2023delivering} achieves multi-modal semantic segmentation with arbitrary-modal complements by relying on the RGB branch and treating other modalities as auxiliary inputs. However, the RGB modality may struggle in certain circumstances, \eg, nighttime, thus it is imperative for the fusion model to learn a more resilient multi-modal framework without relying on a specific sensor.
Our MAGIC framework treats all visual modalities equally and avoids relying on each modality.
This way, our method can eliminate the dependence on RGB modality and better overcome sensor failures while significantly enhancing the segmentation performance.

\section{Methodology}
In this section, we introduce our \textbf{MAGIC} framework. As depicted in Fig.~\ref{overall}, it consists of two pivotal modules: the Multi-modal Aggregation Module (MAM) and the Arbitrary-modal Selection Module (ASM). Our approach takes multiple visual modalities as inputs~\footnote{Here, we take the modalities in DELIVER~\cite{zhang2023delivering} as an example.}. 
\subsection{Task Parameterization} 
\noindent \textbf{Inputs:} Our framework takes input data from four modalities, each captured or synthesized within the same scene. Specifically, we consider RGB images $\textbf{R} \in \mathbb{R}^{h \times w \times 3}$, depth maps $\textbf{D} \in \mathbb{R}^{h \times w \times C^D}$, LiDAR point clouds $\textbf{L} \in \mathbb{R}^{h \times w \times C^L}$, and event stack images $\textbf{E} \in \mathbb{R}^{h \times w \times C^E}$. Here, $C^D=C^L=C^E=3$. Additionally, we incorporate the corresponding ground truth label $\textbf{Y}$ spanning $K$ categories. Unlike existing methods that process multi-modal data individually, our method takes a mini-batch $\{r, d, l, e\}$ containing samples from all modalities, where $r \in \textbf{R}$, $d \in \textbf{D}$, $l \in \textbf{L}$, and $e \in \textbf{E}$.

\begin{figure*}[t!]
    \centering
    \includegraphics[width=\textwidth]{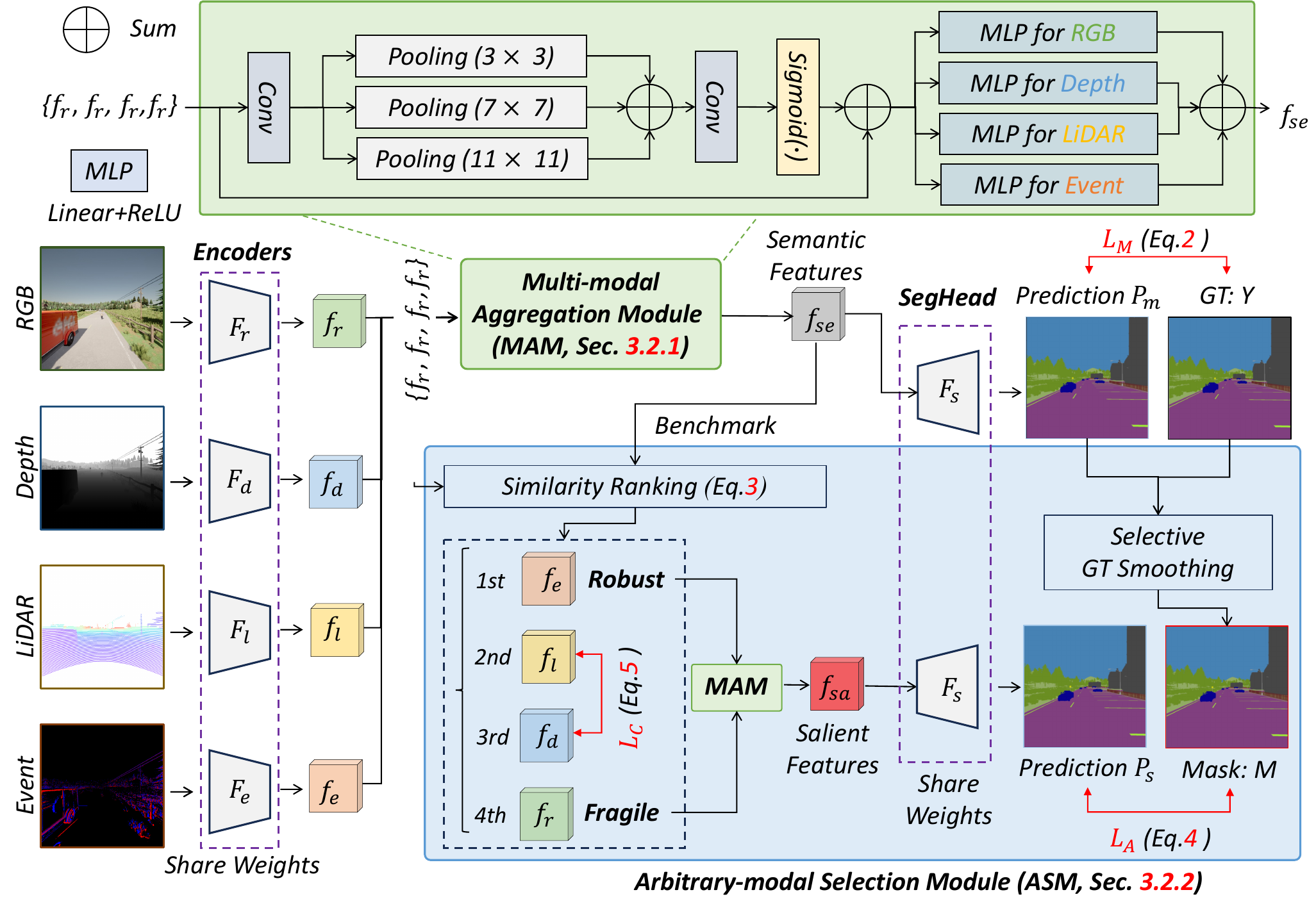}
    \caption{ Overall framework of our MAGIC framework, incorporates plug-and-play multi-modal aggregation and arbitrary-modal selection modules.
    }
    \label{overall}
\end{figure*}

\noindent \textbf{Outputs:} Given the multi-modal data mini-batch $\{r, d, l, e\}$, we feed it into our backbone, producing multi-modal features $\{f_r, f_d, f_l, f_e\}$, as depicted in Fig.~\ref{overall}. Subsequently, these features are concurrently processed by MAM and ASM, yielding semantic feature $f_{se}$ and salient feature $f_{sa}$, respectively. The segmentation head then leverages $f_{se}$ and $f_{sa}$ to derive the MAM predictions $P_{m}$ and the ASM predictions $P_{a}$, respectively.
\subsection{MAGIC Architecture}
As depicted in Fig.~\ref{overall}, our MAGIC framework adopts prevailing backbone models, such as SegFormer~\cite{xie2021segformer}, as the feature encoder and as the segmentation head (SegHead) for each modality. 
The multi-modal mini-batch $\{r,d,l,e\}$ is directly ingested by weight-shared encoders $F_r, F_d, F_l, F_e$ within the backbone model. This process yields high-level multi-modal features $\{f_r, f_d, f_l, f_e\}$ as:
\begin{equation}
\setlength{\abovedisplayskip}{3pt}
\setlength{\belowdisplayskip}{3pt}
\{f_r, f_d, f_l, f_e\} = F_r(r), F_d(d),F_l(l), F_e(e).
\end{equation}
\subsubsection{3.2.1 Multi-modal Aggregation Module (MAM).} 
Upon acquiring the high-level multi-modal features, MAM is designed to further extract the semantic-rich feature from $\{f_r, f_d, f_l, f_e\}$, paving the way for achieving robust arbitrary-modal capabilities. Our MAM centers the values of every modality and simultaneously extracts complementary features from every modality. As shown in Fig.~\ref{overall}, multi-modal data is formed as a batch rather than concatenated and MLP layers are assigned to specific modalities.\textit{ Note that, when performing modality-agnostic validation with arbitrary input modalities, only the input modalities' corresponding layers are used to give predictions.}
This essentially differs from prior methods, \eg,~\cite{zhang2023delivering}, which employs the Self-Query Hub and Parallel Pooling Mixer to prioritize extracted features from auxiliary modalities before fusing them with the RGB feature.
As depicted in Fig.~\ref{overall}, the architecture of MAM comprises three main components: a parallel multi-layer perceptron (MLP), a parallel pooling layer, and integrated residual connections.

Specifically, as shown in Fig.~\ref{overall}, $\{f_r, f_d, f_l, f_e\}$ is first processed with a \textit{Conv} layer, then parallel pooling layers (3$\times$3, 7$\times$7, 11$\times$11) are employed to explore spatial information from different scales. These features are then aggregated and processed with another \textit{Conv} layer with Sigmoid activation.  
Then, parallel MLP layers are utilized to aggregate these multi-modal features with the original $\{f_r, f_d, f_l, f_e\}$.  The aggregated feature from outputs of MLP layers, denoted as $f_{se}$, acts as the semantic representation for generating predictions $P_{m}$ via the segmentation head, \ie, decoder $F_s$. The predicted $P_{m}$ is supervised by the ground truth (GT), $y \in \textbf{Y}$.
Ultimately, the cross-entropy is used as the supervision loss $\mathcal{L}_{M}$ :
\begin{equation}
\setlength{\abovedisplayskip}{3pt}
\setlength{\belowdisplayskip}{3pt}
    \mathcal{L}_{M} = -\sum_{0}^{K-1} \textit{Y}\cdot log(P_{m}).
\end{equation}

\subsubsection{3.2.2 Arbitrary-modal Selection Module (ASM).} 
\begin{wrapfigure}{t}{0.6\textwidth} 
  \centering
  \includegraphics[width=\linewidth]{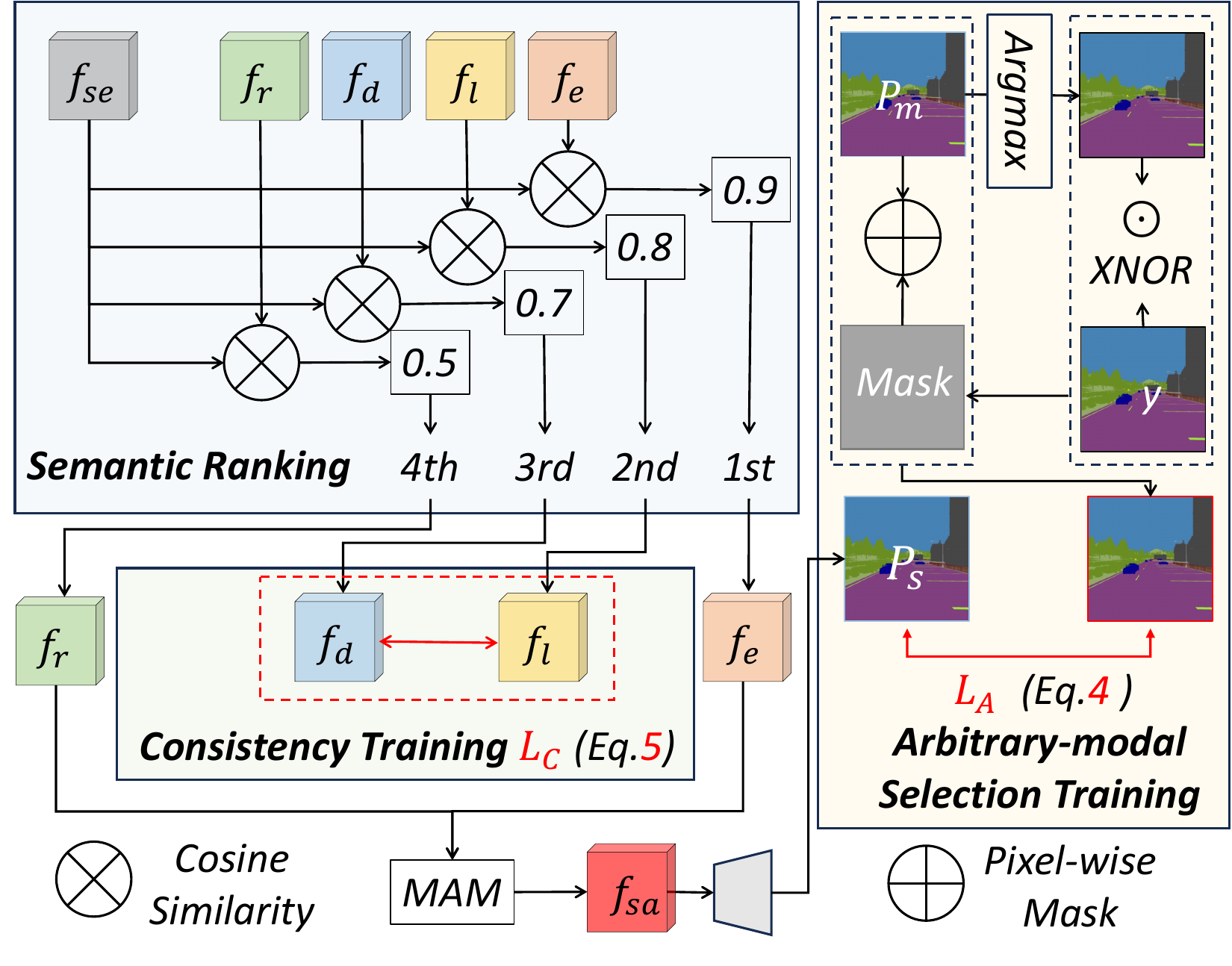}
    \caption{Illustration of the proposed arbitrary-modal selection module (\textbf{ASM}). 
    }
    \label{fig:SAFM}
\end{wrapfigure}

Alongside MAM, we introduce the ASM which is utilized during training, leveraging the most robust feature to \textit{enhance the predictive accuracy} of the framework. 
The integration of the most fragile features — those which are extracted from the challenging input data samples — serves to \textit{reinforce the framework's resilience against missing modalities} in such challenging scenarios. As illustrated in Fig.~\ref{fig:SAFM}, our ASM encompasses two principal components: cross-modal semantic similarity ranking and cross-modal semantic consistency training. We now describe the details. 

\noindent \textbf{  Cross-modal Semantic Similarity Ranking.} serves as a mechanism to compare the multi-modal features $\{f_r, f_d, f_l, f_e\}$ against the semantic feature $f_{se}$ derived from MAM. This thereby yields ranked similarity scores. 
The nature of multi-modal data is inherently diverse, spanning a broad spectrum of conditions. A case in point is the \textit{DELIVER} dataset. As delineated in~\cite{zhang2023delivering}, it features four unique environmental scenarios and documents five episodes of partial sensor malfunctions. Additionally, the intricacies of real-world environments may pose even more heterogeneous challenges. In light of such complexities, it becomes imperative for neural networks to adeptly discern the robust modalities from the fragile ones at the feature level. 
By integrating both the pinnacle of robustness and the nadir of fragility in modalities, a more resilient multi-modal framework can be cultivated. ASM employs the semantic feature $f_{se}$ from MAM as a benchmark to rank $\{f_r, f_d, f_l, f_e\}$ as:
\begin{equation}
\setlength{\abovedisplayskip}{3pt}
    f_{rf}, f_{rm} = \textit{Rank}(\textit{Cos}(\{f_r, f_d, f_l, f_e\}, f_{se})),
\setlength{\belowdisplayskip}{3pt}
\end{equation}
where $f_{rf}$ denotes the features that consist of the top-1 (Robust) and last-1 (Fragile) features in this semantic ranking, signifying the most robust and most fragile modalities, respectively; $f_{rm}$ represents the remaining features in the semantic ranking; $\textit{Rank}$ means sorting from largest to smallest; and $\textit{Cos}(\cdot)$ is the cosine similarity.
The resulting $f_{rf}$ are then passed to another MAM to aggregate the final salient feature for generating predictions $P_{s}$ with the SegHead.

Beyond the scope of feature-level modality-agnostic learning, our ASM, depicted in Fig.~\ref{fig:SAFM}, introduces a prediction-level arbitrary-modal selection training strategy, centered on $y$. It aims to bolster the supervision of predictions $P_{a}$ by selectively merging predictions $P_{m}$ with the hard ground truth label, $y$.
Initially, the logits of the predictions $P_{m}$ undergo an $\textit{argmax}$ operation. These are then amalgamated with the hard label $y$, culminating in the creation of a mask, denoted as $M$. This mask $M$ is designed to retain the predicted logits when category predictions from $argmax(P_{m})$ coincide with those of $y$ for a given pixel. 
Conversely, it discards predicted logits in instances of discrepancy between the category predictions. Thereafter, this mask $M$ is harnessed as the supervision signal for $P_{s}$. This strategy ensures that the arbitrary-feature predictions receive focused and positive supervision and fosters enhanced convergence and averts potential training instabilities.
Overall, the arbitrary-modal selection loss is:
\begin{equation}
\setlength{\abovedisplayskip}{3pt}
\setlength{\belowdisplayskip}{3pt}
    \mathcal{L}_{S} = -\sum_{0}^{K-1} \textit{M}\cdot log(P_{s}).
\end{equation} 
\noindent \textbf{  Cross-modal Semantic Consistency Training.}
With the cross-modal semantic similarity ranking, the top-1 and last-1 ranked features are attained. 
We then \textbf{\textit{impose semantic consistency}} training between the remaining features $f_{rm}$, see Fig.~\ref{fig:SAFM}. The intuition is that the captured semantics of a scene are identical across modalities because the multi-modal data is captured in the same scenario. 
Meanwhile, due to the distinct data formats and unique properties of different sensors, it is non-trivial to directly align the remaining features $f_{rm}$ from different modalities. Intuitively, ASM also takes the semantic feature $f_{sa}$ from MAM as the surrogate and implicitly aligns the correlation, \ie, cosine similarity, between the remaining features and the semantic feature. For brevity, we use the abbreviation $c_1 = \textit{Cos}(f_{rm}^1, f_{sa})$ and $c_2 = \textit{Cos}(f_{rm}^2, f_{sa})$ to represent the correlations. 
The consistency training loss can be formulated as:
\begin{align}
\setlength{\abovedisplayskip}{3pt}
\setlength{\belowdisplayskip}{3pt}
    \mathcal{L}_{C} = \sum_{0}^{K-1} (c_1 log \frac{c_1}{\frac{1}{2} (c_1 + c_2)}
     + c_2 log \frac{c_2}{\frac{1}{2} (c_1 + c_2)}). 
\end{align}
This implicit alignment makes it better to align the features from the scene-semantic consistency perspective. 

\noindent \textbf{Training}
We train our MAGIC framework by minimizing the total loss $\mathcal{L}$ -- a linear combination of the losses of $\mathcal{L}_{M}$,  $\mathcal{L}_{S}$,  and $\mathcal{L}_{C}$:
\begin{equation}
\setlength{\abovedisplayskip}{3pt}
\setlength{\belowdisplayskip}{3pt}
    \mathcal{L} = \mathcal{L}_{M} + \lambda \mathcal{L}_{S} + \beta \mathcal{L}_{C},
\end{equation}
where $\lambda$ and $\beta$ are hyper-parameters for trade-off. The ASM is only utilized in training while the inference is achieved by the backbone together with our MAM.
\begin{figure*}[t!]
    \centering
    \includegraphics[width=0.99\textwidth]{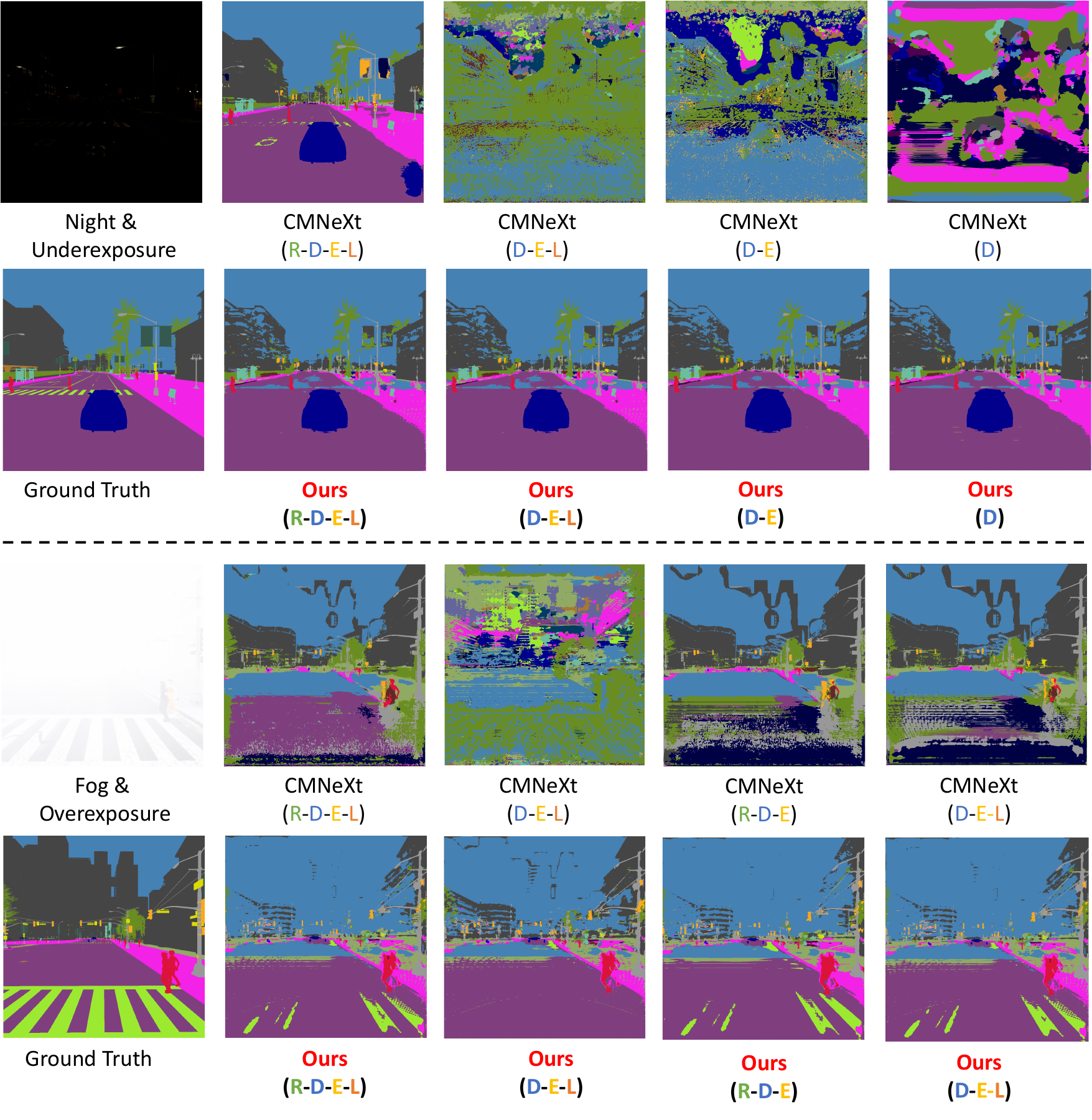}
    \caption{Visualization of arbitrary inputs using \{\textbf{R}GB, \textbf{D}epth, \textbf{E}vent, \textbf{L}iDAR\} on DELIVER.  \textit{(More visualization under different conditions refer to the suppl.)}}
    \label{vis_compare}
\end{figure*}

\section{Experiments} 
\subsection{Datasets and Implementation Details} 
\noindent\textbf{Datasets}
\textbf{1) DELIVER~\cite{zhang2023delivering}} is a large-scale multi-modal dataset that contains depth, LiDAR, Views, Event, and RGB data, with precise annotations of 25 semantic categories. DELIVER considers both environmental conditions (cloudy, foggy, night, and rainy) and sensor failure cases (motion blur, over-exposure, under-exposure, LiDAR-Jitter, \etc) to introduce challenges for robust perception. We follow the official training settings in ~\cite{zhang2023delivering} and convert all data into three channel images.
\textbf{2) MCubeS~\cite{Liang_2022_CVPR}} is a material segmentation dataset with 20 categories and pairs of RGB, Near-Infrared (NIR), Degree of Linear Polarization (DoLP), and Angle of Linear Polarization (AoLP) images.

\noindent \textbf{Implementation Details}
We train our framework on 8 $\times$ NVIDIA GPUs with an initial learning rate of 6$e^{-5}$, which is scheduled by the poly strategy with power 0.9 over 200 epochs. The first 10 epochs are to warm-up framework with 0.1 $\times$ the original learning rate. We use AdamW optimizer with epsilon 1$e^{-8}$, weight decay 1$e^{-2}$, and the batch size is 1 on each GPU. 
The images are augmented by random resize with ratio 0.5-2.0, random horizontal flipping, random color jitter, random gaussian blur, and random cropping to 1024 $\times$ 1024 on DELIVER~\cite{zhang2023delivering}, while to 512 $\times$ 512 on MCubeS~\cite{Liang_2022_CVPR}. ImageNet-1K pre-trained weight is used as the pre-trained weight for the backbone model.

\noindent \textbf{Experimental Settings.} 
\textbf{ 1) Multi-modal Semantic Segmentation} refers to the process of training and validating models using the same modality inputs, without any absence of training modality data.
\textbf{2) Modality-agnostic Semantic Segmentation} means we evaluate all the possible combinations of input modalities and average all results to obtain final mean results.
\begin{table}[t!]
\caption{Per-class results on DELIVER~\cite{zhang2023delivering} and MCubeS~\cite{Liang_2022_CVPR} datasets. Abbreviations: M.: Method; B.b.: Backbone; \#P(M): \#Param(M); Seg-B2: SegFormer-B2. (\textit{Selected categories in this table, all categories refer to the suppl.})}
\resizebox{\linewidth}{!}{
\renewcommand{\tabcolsep}{1pt}
\begin{tabular}{c|c|c|c|ccccccc|c}
\toprule
\multirow{6}{*}{\rotatebox[origin=c]{90}{DELIVER~\cite{zhang2023delivering}}} & Method & Backbone & \#Param (M) & Building & Fence & Pedestrain & Road & Sidewalk & Cars & Wall & \multirow{4}{*}{Mean} \\ \cmidrule{2-11}
 & CMNeXt & Seg-B2 & 58.73 & 89.41 & 43.12 & 76.51 & 98.18 & 82.27 & 84.98 & 69.39 &  \\
 & Ours & Seg-B2 & 24.73 & 89.66 & 49.27 & 76.54 & 98.40 & 86.22 & 90.94 & 70.86 &  \\ \cmidrule{2-11}
 & Method & Backbone & \#Param (M) & Traffic Sign & Ground & Bridge & Groundrail & Static & Water & Terrain &  \\ \cmidrule{2-12}
 & CMNeXt & Seg-B2 & 58.73 & 70.57 & 1.31 & 53.61 & 55.01 & 33.58 & 23.96 & 83.94 & 66.30 \\
 & Ours & Seg-B2 & 24.73 & 72.88 & 2.62 & 59.28 & 73.08 & 35.70 & 30.93 & 84.00 & \textbf{67.66} \\ \bottomrule
\end{tabular}
}
\resizebox{\linewidth}{!}{
\renewcommand{\tabcolsep}{1pt}
\begin{tabular}{c|c|c|c|ccccccc}
\toprule
\multirow{6}{*}{\rotatebox[origin=c]{90}{MCubeS~\cite{Liang_2022_CVPR}}} & Method & Backbone & \#Param (M) & Asphalt & Concrete & Roadmarking & Fabirc & Glass & Plaster & Rubber  \\ \cmidrule{2-11}
 & CMNeXt & Seg-B2 & 58.73 & 84.71 & 45.23 & 74.80 & 32.18 & 54.34 & 0.69 & 28.54  \\ 
 & Ours & Seg-B2 & 24.73 & 88.86 & 50.79 & 74.98 & 38.55 & 55.27 & 0.86 & 34.21   \\ \cmidrule{2-11}
 & Method & Backbone & \#Param (M) & Traffic Sign & Ceramic & Cobblestone & Brick & Water & Sky & Mean   \\ \cmidrule{2-11}
 & CMNeXt & Seg-B2 & 58.73 & 28.72 & 26.81 & 68.67 & 43.19 & 54.35 & 96.52 & 51.54   \\
 & Ours & Seg-B2 & 24.73 & 31.98 & 31.21 & 71.67 & 46.47 & 54.72 & 96.60 & \textbf{53.01} \\ \bottomrule
\end{tabular}}
\label{Tab:Perclass}
\end{table}
\subsection{Experimental Results}
\noindent \textbf{Multi-modal Semantic Segmentation:}
Tab.~\ref{Tab:Perclass} presents a quantitative comparison between our proposed Magic framework and the SoTA, CMNeXt~\cite{zhang2023delivering}, on the DELIVER dataset, utilizing four input modalities. Notably, our Magic framework, which contains only 42\% of the parameters of CMNeXt (specifically \textbf{24.73M} out of 58.73M), surpasses CMNeXt in a majority of categories. 
For instance, for the ‘Fence' category, our model's performance is \textbf{49.27\%}, improving upon CMNeXt's 43.12\% by \textbf{+6.15\%}. Overall, there's a notable increment of +1.33\% in mIoU. 
These results underscore the effectiveness of our multi-modal input batch approach and the proposed plug-and-play modules, positioning them as advancements over the traditional Hub2Fuse, separate branch, and joint branch paradigms (\textit{more comparison refer to the suppl.}). 

In Tab.~\ref{Tab:Perclass}, we present results not only for the DELIVER dataset, which includes RGB, Depth, Event, and LiDAR modalities, but also for the MCubeS~\cite{Liang_2022_CVPR} dataset. The latter comprises four modalities: Image, AoLP, DoLP, and NIR. Notably, our Magic framework consistently surpasses the state-of-the-art CMNeXt~\cite{zhang2023delivering} by an impressive \textbf{+1.47} mIoU. It also achieves superior results in a majority of categories. 
For instance, for the ‘Concrete' category, our model registers \textbf{50.79\%}, outdoing CMNeXt's 45.23\% by \textbf{+5.56\%}. These results emphasize the efficacy of our Magic framework, and they highlight that our proposed MAM and ASM function as efficient plug-and-play modules tailored for multi-modal learning (\textit{For results of training with two modalities, please refer the to suppl.}).
\begin{table}[t!]
\renewcommand{\tabcolsep}{1pt}
\caption{Validation with arbitrary-modal inputs. All methods are trained with four modalities, and the metric is mIoU for all numbers. Abbreviations: Seg-B0: SegFormer-B0, R: RGB; D: Depth; E: Event; L: LiDAR; I: Image; A: Aolp; D: Dolp; N: Nir.}
\resizebox{\linewidth}{!}{
\begin{tabular}{c|c|c|c|cccccccc}
\toprule
\multirow{8}{*}{\rotatebox[origin=c]{90}{DELIVER~\cite{zhang2023delivering}}} & Method & Backbone & \#Param (M) & R & D & E & L & RD & RE & RL & DE \\ \cmidrule{2-12} 
 & CMNeXt & Seg-B2 & 58.73 & 3.76 & 0.81 & \textbf{1.00} & \textbf{0.72} & 50.33 & 13.23 & 18.22 & 21.48 \\ \cmidrule{2-12} 
 & \multirow{2}{*}{\textbf{MAGIC}} & Seg-B0 & \textbf{3.72} & 32.60 & 55.06 & 0.52 & 0.39 & 63.32 & 33.02 & 33.12 & 55.16 \\ \cmidrule{3-12} 
 &  & Seg-B2 & 24.73 & \textbf{41.97} & \textbf{57.59} & 0.40 & 0.37 & \textbf{67.65} & \textbf{41.93} & \textbf{42.00} & \textbf{57.62} \\ \cmidrule{2-12} 
 & Method & Backbone & \#Param (M) & DL & EL & RDE & RDL & REL & DEL & RDEL & Mean \\ \cmidrule{2-12} 
 & CMNeXt & Seg-B2 & 58.73 & 3.83 & \textbf{2.86} & 66.24 & 66.43 & 15.75 & 46.29 & 66.30 & 25.25 \\ \cmidrule{2-12} 
 & \multirow{2}{*}{\textbf{MAGIC}} & Seg-B0 & \textbf{3.72} & 55.17 & 0.26 & 63.37 & 63.36 & 33.32 & 55.26 & 63.40 & 40.49 \tiny{\textbf{+15.24}} \\ \cmidrule{3-12} 
 &  & Seg-B2 & 24.73 & \textbf{57.60} & 0.27 & \textbf{67.66} & \textbf{67.65} & \textbf{41.93} & \textbf{57.63} & \textbf{67.66} & \textbf{44.66} \tiny{\textbf{+19.41}}\\ \midrule
\multirow{8}{*}{\rotatebox[origin=c]{90}{MCubeS~\cite{Liang_2022_CVPR}}} & Method & Backbone & \#Param (M) & I & A & D & N & IA & ID & IN & AD \\ \cmidrule{2-12} 
 & CMNeXt & Seg-B2 & 58.73 & 1.86 & \textbf{1.54} & 2.51 & 2.28 & 47.96 & 43.67 & 45.90 & \textbf{6.99} \\ \cmidrule{2-12} 
 & \multirow{2}{*}{\textbf{MAGIC}} & Seg-B0 & \textbf{3.72} & 46.35 & 0.67 & 33.33 & 1.00 & 47.53 & 47.26 & 45.88 & 0.55 \\ \cmidrule{3-12} 
 &  & Seg-B2 & 24.73 & \textbf{51.91} & 0.32 & \textbf{34.52} & \textbf{2.66} & \textbf{52.24} & \textbf{52.16} & \textbf{52.57} & 1.98 \\ \cmidrule{2-12} 
 & Method & Backbone & \#Param (M) & AN & DN & IAD & IAN & ID+N & ADN & IADN & Mean \\ \cmidrule{2-12} 
 & CMNeXt & Seg-B2 & 58.73 & 7.58 & 9.95 & 50.07 & 48.77 & 48.83 & 8.06 & 51.54 & 25.17 \\ \cmidrule{2-12} 
 & \multirow{2}{*}{\textbf{MAGIC}} & Seg-B0 & \textbf{3.72} & 35.32 & 35.70 & 47.85 & 47.01 & 46.71 & 35.64 & 47.58 & 34.56 \tiny{\textbf{+9.39}} \\ \cmidrule{3-12} 
 &  & Seg-B2 & 24.73 & \textbf{36.01} & \textbf{37.09} & \textbf{52.48} & \textbf{52.82} & \textbf{52.73} & \textbf{37.57} & \textbf{53.01} & \textbf{38.00} \tiny{\textbf{+14.97}}\\ \bottomrule
\end{tabular}
}
\label{Tab:ArbitrarySeg_4modal}
\end{table}

\noindent \textbf{Modality-agnostic Semantic Segmentation}
Unlike the approach in ~\cite{zhang2023delivering} where arbitrary modality inputs necessitate the inclusion of RGB data, our methodology operates effectively on arbitrary inputs without specifically relying on any given modality. To evaluate the resilience of our approach with such arbitrary modality inputs, we apply our Magic framework to both the DELIVER and MCubeS~\cite{Liang_2022_CVPR} datasets.
As shown in Tab.~\ref{Tab:ArbitrarySeg_4modal}, models are trained with the total four modalities and is validated with arbitrary modality combination. Our Magic with SegFormer-B2 significantly outperforms the CMNeXt~\cite{zhang2023delivering} at nearly all the validation scenarios and achieves +19.41\% and +14.97\% mIoU performance gain than CMNeXt~\cite{zhang2023delivering} at DELIVER~\cite{zhang2023delivering} and MCubeS~\cite{Liang_2022_CVPR} datasets, respectively. Especially for the RGB data absence scenarios, our Magic with SegFormer-B2 eclipses much more performance, such as Depth only (\textbf{57.59\%} vs. 0.81\% $\rightarrow$ \textbf{+56.78\%}), Depth with Event (\textbf{57.62\%} vs. 21.48\% $\rightarrow$ \textbf{+36.14\%}), and Depth with LiDAR (\textbf{57.60\%} vs. 3.83 $\rightarrow$ \textbf{+53.77\%}). Importantly, our Magic with SegFormer-B2 only has 42\% parameters of CMNeXt. Furthermore, our Magic with SegFormer-B0 even surpasses CMNeXt~\cite{zhang2023delivering} by \textbf{+15.24} and \textbf{+14.97} mIoU with only \textbf{6\%} parameters. All these results show that our MAM and ASM are powerful plug-and-play modules for multi-modal visual learning, especially for the modality-agnostic segmentation scenarios with arbitrary-modal inputs.
\begin{table}[t!]
\renewcommand{\tabcolsep}{1pt}
\caption{Results of modality-agnostic validation with three modalities.}
\resizebox{\linewidth}{!}{
\begin{tabular}{c|c|c|ccccccc|c|c}
\midrule
\multirow{2}{*}{Method} & \multirow{2}{*}{Backbone} & \multirow{2}{*}{Training} & \multicolumn{7}{c}{DELIVER dataset} & \multirow{2}{*}{Mean} & \multirow{2}{*}{$\Delta \uparrow$} \\ \cmidrule{4-10}
 &  &  & R & D & L & RD & RL & DL & RDL &  &  \\ \midrule
CMNeXt~\cite{zhang2023delivering} & Seg-B2 & \multirow{3}{*}{RDL} & 1.87 & 1.87 & 2.01 & 52.90 & 23.35 & 4.67 & 65.50 & 21.74 & - \\ \cmidrule{1-2} \cmidrule{4-12} 
\multirow{2}{*}{MAGIC(ours)} & Seg-B0 &  & 32.41 & 56.20 & 1.40 & 62.64 & 32.61 & \textbf{56.29} & 62.64 & 43.46 &+21.72 \\ \cmidrule{2-2} \cmidrule{4-12} 
 & Seg-B2 &  & \textbf{37.08} & \textbf{60.52} & \textbf{2.38} & \textbf{67.66} & \textbf{67.62} & 37.36 & \textbf{67.63} & \textbf{54.18} & \textbf{ +32.44} \\ \midrule
\multicolumn{3}{c}{} & R & D & E & RD & RE & DE & RDE & Mean & $\Delta \uparrow$ \\ \midrule
CMNeXt~\cite{zhang2023delivering} & Seg-B2 & \multirow{3}{*}{RDE} & 1.75 & 1.71 & 2.06 & 53.68 & 9.66 & 2.84 & 64.44 & 19.45 & - \\ \cmidrule{1-2} \cmidrule{4-12} 
\multirow{2}{*}{MAGIC(ours)} & Seg-B0 &  & 32.96 & 55.90 & 2.15 & 62.52 & 33.25 & 56.00 & 62.49 & 43.61 & +24.16 \\ \cmidrule{2-2} \cmidrule{4-12} 
 & Seg-B2 &  & \textbf{38.13} & \textbf{60.42} & \textbf{2.75} & \textbf{67.16} & \textbf{38.12} & \textbf{60.87} & \textbf{67.16} & \textbf{47.80} & \textbf{+28.35} \\ \midrule
\multirow{2}{*}{Method} & \multirow{2}{*}{Backbone} & \multirow{2}{*}{Training} & \multicolumn{7}{c}{McubeS dataset} & \multirow{2}{*}{Mean} & \multirow{2}{*}{$\Delta \uparrow$} \\ \cmidrule{4-10}
 &  &  & I & A & N & IA & IN & AN & IAN &  &  \\ \midrule
CMNeXt~\cite{zhang2023delivering} & Seg-B2 & \multirow{3}{*}{IAN} & 2.16 & 2.09 & 2.45 & 41.46 & 44.66 & 10.73 & 47.96 & 21.64 & - \\ \cmidrule{1-2} \cmidrule{4-12} 
\multirow{2}{*}{MAGIC(ours)} & Seg-B0 &  & 47.04 & 0.33 & 33.00 & 47.70 & 46.17 & 34.92 & 47.43 & 36.66 & +15.02 \\ \cmidrule{2-2} \cmidrule{4-12} 
 & Seg-B2 &  & \textbf{50.36} & \textbf{31.25} & \textbf{43.69} & \textbf{50.79} & \textbf{50.35} &\textbf{ 45.92} &\textbf{ 50.80} & \textbf{46.17} & \textbf{+24.53} \\ \midrule
 \multicolumn{3}{c}{} & I & A & D & IA & ID & AD & IAD & Mean & $\Delta \uparrow$ \\ \midrule
 CMNeXt~\cite{zhang2023delivering} & Seg-B2 & \multirow{3}{*}{IAD} & 1.15 & 1.87 & 1.27 & 47.14 & 47.80 & 12.52 & 49.48 & 23.03 & - \\ \cmidrule{1-2} \cmidrule{4-12} 
\multirow{2}{*}{MAGIC(ours)} & Seg-B0 & & 46.58 & 0.01 & 4.84 & 46.68 & 47.51 & 3.87 & 47.78 & 28.18 &  +5.15 \\ \cmidrule{2-2} \cmidrule{4-12} 
 & Seg-B2 & & \textbf{49.05} & \textbf{35.45} & \textbf{39.14} & \textbf{50.52} & \textbf{50.20} & \textbf{41.60} & \textbf{50.79} & \textbf{45.25} & \textbf{+22.22} \\ \bottomrule
\end{tabular}}
\label{Tab:ArbitrarySeg_3modal}
\end{table}

In Tab.~\ref{Tab:ArbitrarySeg_3modal}, we show the modality-agnostic validation results of training models with 3 modalities on both datasets. Our Magic consistently outperforms CMNeXt~\cite{zhang2023delivering} at all arbitrary modality inputs and achieve more performance gains. For instance, our Magic with SegFormer-B2 achieves \textbf{+32.44\%} and \textbf{+28.35\%} mIoU performance gains on DELIVER~\cite{zhang2023delivering}, \textbf{+24.53\%} and \textbf{+22.22\%} mIoU performance boost on MCubeS~\cite{Liang_2022_CVPR} with only \textbf{42\%} parameters. This further confirms the superiority and robustness of our proposed Magic framework in the arbitrary input setting.
In Fig.~\ref{vis_compare}, we present a comparison of the segmentation results obtained using Magic and CMNeXt~\cite{zhang2023delivering}. The results demonstrate that our Magic consistently performs well with arbitrary inputs, whereas CMNeXt is fragile in most scenarios. Notably, our Magic does not rely on a specific modality and is relatively insensitive to the absence of modalities, which further enhances the robustness of full scene segmentation under varying lighting and weather conditions, such as cloudy, rain, and motion blur.

\noindent \textbf{Effectiveness of the Loss Functions.}
To evaluate the effectiveness of the proposed loss functions, we conduct ablation experiments on two datasets, DELIVER and MCubeS, using SegFormer-B0 and -B2 backbones. As delineated in Tab.~\ref{AB:LossFunc}, each of our proposed modules and associated loss functions consistently enhances the performance of multi-modal semantic segmentation. Significantly, our MAM in conjunction with $\mathcal{L}_M$ results in mIoU improvements of \textbf{+8.21\%} and \textbf{+5.51\%} for SegFormer-B0 and -B2, respectively. Building upon our MAM, the ASM and its associated $\mathcal{L}_A$ further register mIoU increments of \textbf{+10.78\%} and \textbf{+8.12\%} across the dual backbones. Complementing these modules, the consistency training loss $\mathcal{L}_C$ yields \textbf{+11.99\%} and \textbf{+8.99\%} mIoU advancements relative to the baseline. 

\noindent \textbf{Ablation Study of MAM Components.} 
As indicated in Fig.~\ref{ab_fig} (c), we evaluate individual components within our proposed MAM. Removing any of the components invariably results in diminished performance. 
This underscores the indispensable contribution of each component to the efficacy of MAM.

\noindent \textbf{Ablation of Pooling Size.}
We conduct an ablation study focusing on the pooling size used in the parallel pooling of MAM on MCubeS, as depicted in Fig.~\ref{ab_fig} (d). Our empirical findings indicate that a pooling size of {(3,7,11)} consistently delivers the optimal mIoU performance. \\
\noindent \textbf{Selection of the Salient Features in ASM.}
The process of selecting salient features is pivotal for ASM.  
In order to evaluate the efficacy of this selection strategy, we organize the features based on their congruence with the semantic features, subsequently marking the salient features with stars, as shown in Fig.~\ref{ab_salient} (a). Our results underscore that choosing features from both extremes of the sorted sequence optimally supports modality-agnostic segmentation, resonating with our initial hypothesis. We also conduct experiments by randomly dropping one or a few modalities with only MAM at train time, the result is 42.44 mIoU which is less than ours 44.66 mIoU.
\begin{table}[t!]
\renewcommand{\tabcolsep}{6pt}
\caption{Ablation study of different loss function combinations.}
\resizebox{\linewidth}{!}{
\begin{tabular}{c|ccc|cc|cc|cc|cc}
\midrule
\multirow{3}{*}{Backbone} & \multicolumn{3}{c|}{\multirow{2}{*}{Loss}} & \multicolumn{4}{c|}{\multirow{2}{*}{DELIVER~\cite{zhang2023delivering}}} & \multicolumn{4}{c}{\multirow{2}{*}{MCubeS~\cite{Liang_2022_CVPR}|}} \\
 & \multicolumn{3}{c|}{} & \multicolumn{4}{c|}{} & \multicolumn{4}{c}{} \\ \cmidrule{2-12} 
& $\mathcal{L}_{M}$ & $\mathcal{L}_{S}$ & $\mathcal{L}_{C}$ & mIoU & $\Delta \uparrow$ & Acc & $\Delta \uparrow$ & mIoU & $\Delta \uparrow$ & Acc & $\Delta \uparrow$ \\ \midrule
\multirow{4}{*}{\small{Seg-B0}} & - & - & - & 51.41 & - & 59.76 & - & 42.43 & - & 51.34 & -  \\ 
& \Checkmark & - & - & 59.62 & +8.21 & 68.42 & +8.66 & 42.91 & +0.48 & 52.72 & +1.38 \\ 
 & \Checkmark & \Checkmark & - & 62.19 & +10.78 & 70.12 & +10.36 & 46.50 & +4.07 & 56.11 & +4.77 \\
 & \Checkmark & \Checkmark & \Checkmark & 63.40 & +11.99 & 70.84 & +11.08 & 47.58 & +5.15 & 56.35 & +5.01 \\ \midrule
\multirow{4}{*}{\small{Seg-B2}} & - & - & - & 58.67 & - & 65.48 & - & 46.51 & - & 56.02 & - \\ 
& \Checkmark & - & - & 64.18 & +5.51 & 73.35 & +7.87 & 49.66 & +1.80 & 58.67 & +1.03 \\ 
 & \Checkmark & \Checkmark & - & 66.79 & +8.12 & 74.08 & +8.60 & 52.24 & +4.38 & 61.70 & +4.06 \\ 
 & \Checkmark & \Checkmark & \Checkmark & \multicolumn{1}{l}{67.66} & +8.99 & \multicolumn{1}{l}{74.69} & +9.21 & \multicolumn{1}{l}{53.01} & +5.15 & \multicolumn{1}{l}{62.87} & +5.23 \\ \bottomrule
\end{tabular}}
\label{AB:LossFunc}
\end{table}

\section{Ablation Study and Analysis}
\begin{figure}[t!]
    \centering
    \includegraphics[width=\linewidth]{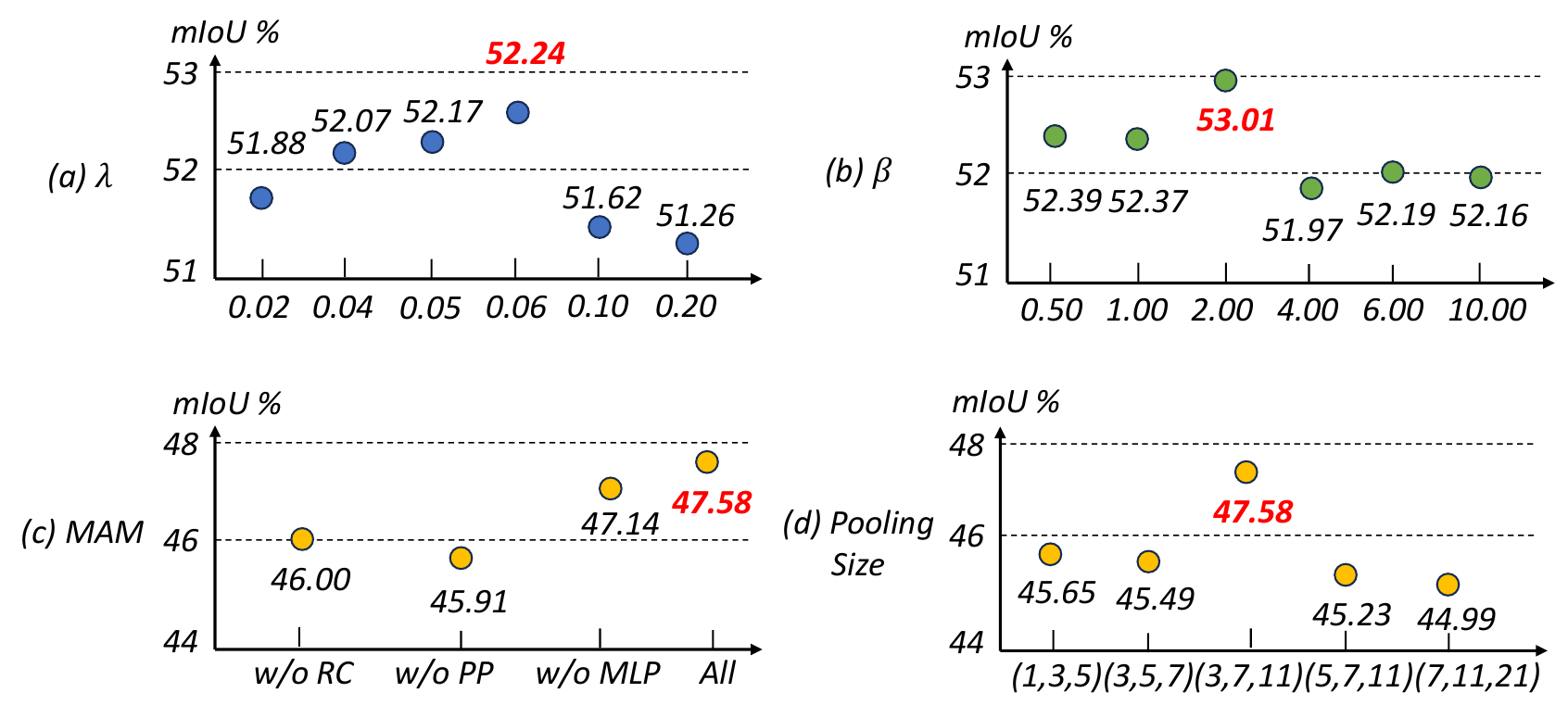}
    \caption{(a),(b): Ablation of $\lambda$ and $\beta$ on MCubeS~\cite{Liang_2022_CVPR}. (c),(d): Ablation of components in MAM with MiT-B0 on MCubeS~\cite{Liang_2022_CVPR}.}
    \label{ab_fig}
\end{figure}
\noindent \textbf{Ablations of $\lambda$ and $\beta$.}
Fig.~\ref{ab_fig} (a) and (b) present the ablation results for varying values of hyper-parameters $\lambda$ and $\beta$.

\section{Discussion}
\noindent \textbf{Every Modality Matters.}
Contrary to the assumptions in prior works such as~\cite{zhang2023delivering}, which advocate for the indispensability of RGB representation in semantic segmentation, we propose that every modality holds significance. Our contention is that every modality brings value and should not be overlooked. As evidenced in Tab.~\ref{Tab:ArbitrarySeg_4modal}, introducing each modality during inference within our Magic model corresponds to enhancements in mIoU; for instance, transitioning from R+D+L to R+D+E+L improves from 67.65\% to 67.66\%. 
\begin{figure}[t!]
    \centering\includegraphics[width=\linewidth]{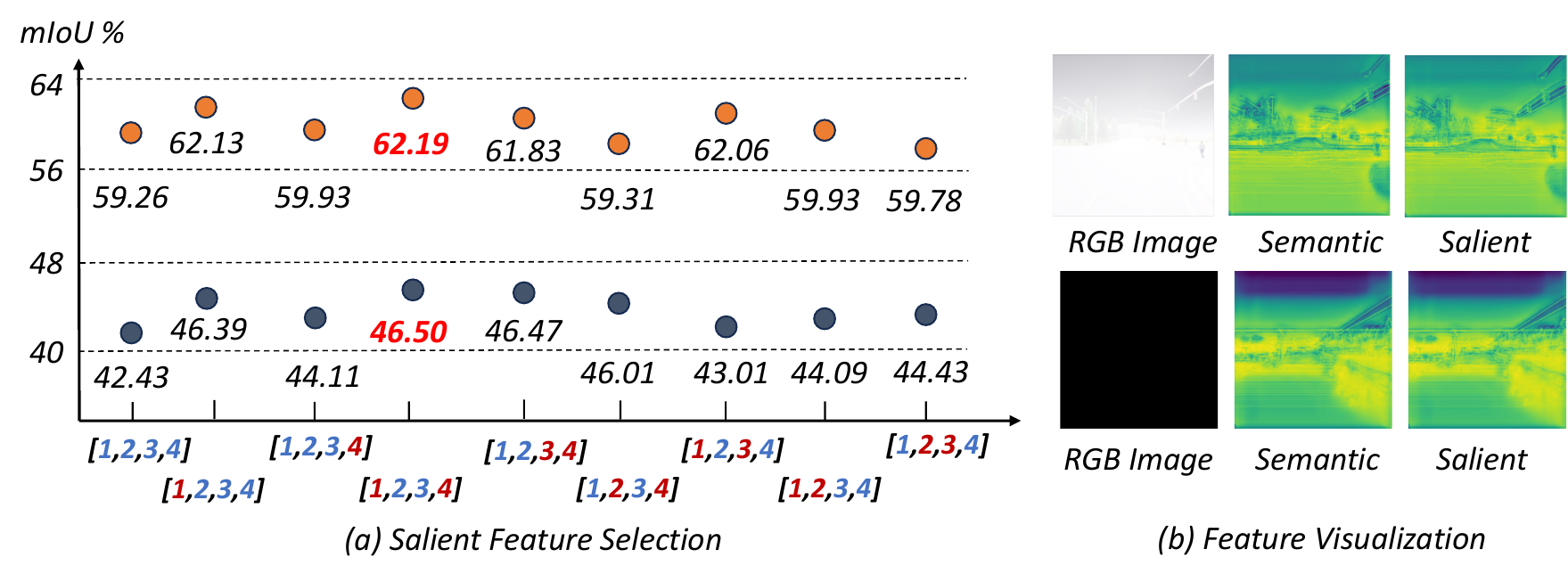}
    \caption{(a) Ablation study on the salient feature selection strategy, where [1,2,3,4] stand for the ranked multi-modal features and the \textbf{\textcolor{blue}{blue}} numbers denotes unselected features, and \textcolor{red}{\textbf{red}} numbers represents the selected salient features among the ranked multi-modal features. (b) Visualization of both semantic and salient features.}
    \label{ab_salient}
\end{figure}

\noindent \textbf{Visualization of Salient Features.}
In Fig.~\ref{ab_salient} (b), we provide visualization of the RGB features, semantic features derived from MAM, and the salient features extracted by ASM. Notably, both the semantic and salient features exhibit a richer capture of scene details than the RGB features. This underscores the efficacy of our MAM and ASM in harnessing the potential of multi-modal input.


\section{Conclusion}
In this paper, we presented our proposed Magic framework for modality-agnostic semantic segmentation, which can be implemented with various existing segmentation backbones. We introduced a multi-modal aggregation module (MAM) for extracting complementary scene information and a unified arbitrary-modal selection module (ASM) for enhancing the backbone model's robustness for arbitrary-modal input data. Our Magic significantly outperformed previous multi-modal methods in both multi-modal and modality-agnostic semantic segmentation benchmarks by a large margin, achieving new state-of-the-art performance.

\noindent \textbf{Limitations and Future Directions.} Despite its notable contributions, MAGIC exhibits sub-optimal performance due to inherent data characteristics in a few scenarios. Future efforts will focus on refining and enhancing our modality-agnostic plug-and-play modules to ensure consistent and improved performance.

\clearpage
\section{Acknowledgement}
This paper is supported by the National Natural Science Foundation of China (NSF) under Grant No. NSFC22FYT45, the Guangzhou City, University and Enterprise Joint Fund under Grant No.SL2022A03J01278, and Guangzhou Fundamental and Applied Basic Research (Grant Number: 2024A04J4072)

%
\bibliographystyle{splncs04}
\bibliography{main}
\end{document}